# Algorithmic Bias from the Perspectives of Healthcare Professionals

Jennifer Xu[1][a] and Tamara Babaian[1][b]
[1]*Department of Computer Information Systems, Bentley University, Waltham, MA, U.S.A.*
*{jxu, tbabaian}@bentley.edu*

Abstract: This paper focuses on algorithmic bias of machine learning and artificial intelligence applications in healthcare information systems. Based on the quantitative data and qualitative comments from a survey and interviews with healthcare professionals, who have different job roles (e.g., clinical vs. administrative), this study provides findings about the relationships between algorithmic bias, perceived fairness, and the intended acceptance and adoption of ML algorithms and algorithm generated outcomes. The results suggest that the opinions of healthcare professionals toward the causes of algorithmic bias, the criteria of algorithm assessment, the perceived fairness, and bias mitigation approaches may vary depending on their job roles, perspectives, tasks, and the algorithm characteristics. More research is needed to investigate algorithmic bias to ensure fairness and equality in healthcare.

## 1 INTRODUCTION

Artificial intelligence (AI) has been increasingly used and adopted by individuals, organizations, and institutions around the globe. AI can be employed to support decision making and enhance productivity in a broad spectrum of application domains, such as business, finance, transportation, and education. AI and machine learning (ML) algorithms have also been used in the medical and healthcare domain for clinical decision support (Rajkomar et al., 2018), such as predicting hypertension and obesity (Ge et al., 2023; Gupta et al., 2022), diagnosing cardiovascular diseases (Litjens et al., 2019), detecting cancerous tumors (Lehman et al., 2015), identifying high-risk, high-cost patients (Osawa et al., 2020), and optimizing clinical workflow (Akkus, 2021).

While AI and ML have found many promising applications in healthcare, many healthcare stakeholders (e.g., physicians, hospital managers, payers, and patients) have been increasingly concerned with algorithmic bias, which may cause serious consequences for clinical safety (Challen et al., 2019) and misalignments with ethical principles (Morley et al., 2020). A recent comprehensive study of clinical ML algorithms shows that in several medical disciplines, such as cardiology, nephrology, and obstetrics, using patient race and ethnicity in ML algorithms may lead to the conclusion that Black patients are in less need for care (Vyas et al., 2020).

There have been many proposals for addressing ML algorithmic bias and related ethical issues, such as fairness, equality, discrimination, among many others (Mehrabi et al., 2021). However, the problem remains challenging to tackle, especially in healthcare, due to several reasons.

First, it can be difficult to identify the sources of algorithmic bias. There are a wide variety of biases, which may originate from different sources and caused by different reasons (Giovanola & Tiribelli, 2023; Kordzadeh & Ghasemaghaei, 2022). For example, biased outcomes produced by an ML algorithm may be caused by human bias embedded in the training data (Gaonkar et al., 2020; Larrazabal et al., 2020). Algorithms trained on data from one community may be biased when utilized in another community with a different patient population (Liu et al., 2018). Rarity of certain medical conditions and lack of clinical expertise may also result in imbalanced samples, leading to biased, unfair outcomes (Ktena et al., 2024).

Second, since ML algorithms are used to support decision making, their performance (e.g., accuracy, sensitivity) is one of the most important criteria for algorithm assessment. However, it can be practically impossible for ML algorithms to satisfy all performance criteria and ethical principles at the same time (Giovanola & Tiribelli, 2023). For example, as fairness is a perception (Kordzadeh & Ghasemaghaei, 2022; Wang et al., 2020), what one social group perceives as fair may be considered unfair by other groups (Ochmann et al., 2024). In addition, the variety in algorithms adds more complexity to the task of assessing and selecting algorithms when facing algorithmic bias. There are many types of ML algorithms, ranging from supervised learning, unsupervised learning, to generative AI algorithms. Even within the same learning category, algorithms may have heterogeneous characteristics in their design, architecture, and parameter setting. For instance, although both decision tree algorithms and neural networks can be used in supervised learning to classify data, their design, inner workings, and learned models are completely different. The classifiers learned by decision trees are often easy to interpret and explain, yet neural networks are considered "black box," lacking transparency and explainability.

[a] https://orcid.org/ 0000-0001-5615-9967
[b] https://orcid.org/0009-0000-4341-042X

Third, it is unclear how different healthcare stakeholders view algorithmic bias and mitigation strategies. Different stakeholders may have different attitudes and opinions toward many questions, such as the presence of algorithmic bias in healthcare ML algorithms, where the biases come from, how to mitigate the biases, and which algorithms to select and adopt (Vorisek et al., 2023). Their opinions may depend on several factors, such as their roles and perspectives, the tasks they wish to use AI and ML to accomplish, and the priority of goals. For example, doctors and physicians may focus on accuracy of diagnoses and effectiveness of treatments; managers may prioritize patient safety and hospital operational efficiency over other aspects; patients may wish to lower medical costs in addition to receiving timely, quality care.

This research seeks to study the opinions of healthcare professionals toward algorithmic bias and fairness. We intend to explore the following research questions (RQs):

- RQ1: What do healthcare professionals think about the causes of algorithmic bias?
- RQ2: What approaches do they believe would help address bias and fairness?
- RQ3: What criteria and factors do they consider when selecting algorithms?
- RQ4: How would they intend to use and adopt ML algorithms in their practice and work?

Using a survey and interviews, we gathered quantitative data and qualitative comments from healthcare professionals from different hospitals. The participants have various job roles in their hospitals ranging from clinical (e.g., physicians, doctors, and nurses), technical (e.g., radiologists), administrative, to IT (e.g., information system developers) and support (e.g., trainers and educators). Our findings show that healthcare professionals are concerned about algorithmic bias and fairness, and that their opinions about using AI and ML algorithms in their practice of medicine and healthcare management differ depending on several factors.

The remainder of the paper is organized as follows. The next section reviews the literature about algorithmic bias, the application of ML algorithms and bias identified in the literature. Section 3 describes research methods and data, followed by reports of analysis and results in Section 4. Section 5 discusses our findings and implications. The last section outlines plans for future research and concludes the paper.

# 2 LITERATURE REVIEW

## 2.1 Algorithmic Bias

The concept of computer system bias is not new (Moor, 1985); it was further developed by Friedman & Nissenbaum (Friedman & Nissenbaum, 1996) and has since evolved into what is now commonly referred to as algorithmic bias, particularly, within the scope of AI/ML systems. Algorithmic bias refers to systematic errors which may disadvantage specific individuals or groups of population without a justified reason (Kordzadeh & Ghasemaghaei, 2022). Algorithmic bias is a socio-technical concept: it is rooted in the biases that exist in society, which make their way into technology, the use of which, in-turn, may contribute to helping proliferate and amplify discriminatory practices by humans.

Within the context of ML applications, algorithmic bias can be a result of inappropriate choice of training data, model, or inappropriate use of a system (Giovanola & Tiribelli, 2023), categorized, respectively, as data-driven bias, model design bias, and user-interaction bias. Data-driven biases originate from inadequate representation (minority or selection bias), missing data, and differences between the population in the training and deployment data (domain shift bias). Model bias arises during the model conceptualization stages, for example, with the selection and assignment of classification labels (Zając et al., 2023). Biases arising from the interaction of users and ML technology (sometimes referred to as latent (DeCamp & Lindvall, 2020), or emergent (Friedman & Nissenbaum, 1996) bias, include automation and feedback loop biases, which are caused by overreliance on the potentially imperfect decision support algorithms without thorough questioning of the system-generated predictions.

Fairness, one of the core ethical principles, is closely related to algorithmic bias. In the healthcare context, unfair outcomes often are related to the use of protected attributes, such as gender, age, race, ethnicity, socioeconomic status, etc. (Abràmoff et al., 2023). As an ethics value, fairness has multiple dimensions including distributive, procedural, interpersonal, and informational justice (Ochmann et al., 2024). In our research, we focus on the *perceived fairness*, which is defined as the extent to which algorithms are perceived to be fair by people (Kordzadeh & Ghasemaghaei, 2022).

While ML models are typically developed and assessed with the goal of optimizing for specific overall performance measures, such as accuracy, precision, and sensitivity, it remains a challenge, in general, for ML algorithms to achieve high performance while aligning with all ethical principles at the same time (Giovanola & Tiribelli, 2023).

## 2.2 Algorithmic Bias in Healthcare

A family of decision-tree-based algorithms have been employed in clinical and healthcare management applications and research, such as predicting survival in locally advanced rectal cancer (De Felice et al., 2020), readmission in mental health patients (Morel et al., 2020), and triage level designation for emergency room (ER) patients (Levin et al., 2018). Deep learning algorithms that use neural networks have also been employed to process medical imagery data (e.g., X-ray, MRI, CT scans, and ultrasound images) for detecting, screening, or analyzing various clinical conditions, such as breast tumors (Lehman et al., 2015), lung cancers (Ardila et al., 2019), and cardiovascular complications (Litjens et al., 2019), and to classify numerical and text data (e.g., visual signs and clinical notes) for predicting hypertension (Ge et al., 2023), diagnosing cancers (Fu et al., 2020), preventing inpatient falls (Cheligeer et al., 2024), and providing new disease insights (Rajpurkar et al., 2022).

While ML algorithms are employed in healthcare, there have been growing concerns about algorithmic bias that may jeopardize clinical safety (Challen et al., 2019) and cause healthcare inequality, disparity, and unfair outcomes toward underrepresented social groups (Giovanola & Tiribelli, 2023; Mehrabi et al., 2021; Schrouff et al., 2023).

Investigations of the sources and types of algorithmic bias in clinical decision support applications confirm that algorithmic bias in healthcare may originate from the training data, the algorithm design, or the interactions between human users (e.g., physicians and patients) and clinical support systems (Giovanola & Tiribelli, 2023). The distribution shift, which occurs when there are discrepancies between the training data and the data in real-world settings, can cause the learned model to perform and generalize poorly and produce biased outcomes. For example, a study finds that a large performance drop occurs when an ML model trained on data from 17 teledermatology services in the U.S. is applied to teledermatology cases in Colombia (Schrouff et al., 2023). Similarly, Watson for Oncology, a system with ML algorithms trained on Western datasets, is found to perform much worse when used for Chinese patients (Liu et al., 2018). Imbalanced data caused by underrepresentation of some social groups (e.g., gender) may lead to biased classifiers (Larrazabal et al., 2020). Labelling bias resulted from subjective annotations of physicians or use of billing and reimbursement driven diagnostic coding of diseases may also cause a trained model to reflect the bias embedded in the training data (Yu & Kohane, 2019). Automation bias (a.k.a., confirmatory bias) may occur when physicians over rely on algorithm generated recommendations and diagnoses (Giovanola & Tiribelli, 2023). For example, automation bias is shown to cause an increased false negative rate in radiology diagnoses (Lehman et al., 2015).

The discussion of fairness in healthcare often centers on distributive fairness (Giovanola & Tiribelli, 2023). A recent study shows that including patient race and ethnicity information in the data for ML algorithms may potentially lead to unfair distribution of clinical resources toward certain minority groups (Vyas et al., 2020). However, it has also been reported that in some cases, ML algorithms may perform well with fair outcomes for different social groups (Noseworthy et al., 2020) and across hospitals and sections (Levin et al., 2018).

Researchers propose frameworks and guidelines for building safer ML supported clinical decision systems (Challen et al., 2019), and promoting trust (Ema et al., 2020). However, development of strategies and methods for mitigating bias and enhancing fairness and trust in ML algorithms in healthcare remains elusive, due to the complexity of the issues. It is widely recognized that the demands on effectiveness, safety, and fairness of AI-based tools require that healthcare AI developers, users, and regulating bodies work collaboratively to define guidelines for clarifying the levels of transparency on the provenance, quality of data as well as assessment mechanisms for the AI tools before their adoption (Matheny et al., 2023). Furthermore, the risks associated with the entrenchment and amplification of existing biases due to the use of black-box decision support require specific attention to the clinicians' understanding of the potential risks to patient safety and equity, which motivates our study.

### 2.3 Stakeholder Perspectives

Healthcare is an industry with many different stakeholders, including healthcare organizations, physicians, administrative and support personnel, payers and insurance companies, clinical information system and electronic health record (EHR) developers, and patients. Different stakeholders may have diverse perspectives, priorities, and opinions regarding ML algorithms, algorithmic bias, and fairness.

There has been limited research and reports on stakeholder perspectives. A qualitative study (Parikh et al., 2022) reports findings from interviews of 29 oncology clinicians regarding their perceptions on the adoption of ML-based predictions of patient mortality risk to prompt conversations of end-of -life care with cancer patients. It is found that physicians are generally positive toward the prospect of using ML generated predictions. However, they are concerned with ethical issues, accuracy and possible confirmation and automation biases. Another study has conducted interviews with a group of healthcare AI experts specialized in system development and regulation. The findings show that experts' opinions vary regarding the mitigation strategies for algorithmic bias (Aquino et al., 2023). Specifically, there are divergent views about whether protected attributes such as sociocultural identifiers (e.g., race and gender) should be included in healthcare ML algorithms (US Department of Health and Human Services 2024). Similarly, a web-based survey has found that healthcare AI developers perceive their algorithms to be moderately fair (Vorisek et al., 2023).

Our research seeks to explore how healthcare professionals view and perceive algorithmic bias and fairness and how they intend to assess, select, and adopt ML algorithms when facing algorithmic bias. Our chosen theoretical framework presented in (Kordzadeh & Ghasemaghaei, 2022) focuses on algorithmic bias, perceived fairness, and user behavioral responses. It posits that algorithmic bias negatively influences perceived fairness, which further affects the behavioral response of users in terms of their acceptance of the algorithms and adoption of ML based decision support systems. Moreover, these relationships are affected by several factors including individual, task, and technology characteristics. For example, individuals with different education levels and gender may perceive the outcome of ML algorithms differently (Wang et al., 2020), and the responses to algorithmic bias may also vary depending on whether the task has high-impact or low-impact. Technology characteristics, such as levels of transparency and explainability, may also affect the perceptions of fairness, and users' behavioral responses.

## 3 METHODS AND DATA

Our research methodology includes survey and interviews. Both survey and interviews target healthcare professionals taking various roles in their organizations.

### 3.1 Survey

The survey is designed to explore opinions and attitudes of healthcare professionals toward algorithmic bias, fairness, and intended behavior in response to bias. To reduce the scope of ML algorithms referred to in the survey, we focus only on supervised ML classification algorithms.

The survey consists of two parts. The first part contains eight questions regarding the participant demographic information (e.g., age, gender), background (e.g., job roles in organizations, years of work experience, and knowledge about ML algorithms), and their general attitudes toward AI technology in general. The second part includes five questions using a five-point Likert scale ranging from Strongly Disagree to Strongly Agree. Each of these questions focuses on the participant's opinions toward a specific topic: causes of algorithmic bias (Q9), fairness (Q10), algorithm assessment and task characteristics (Q11), technology characteristics (Q12), and intended behavior (acceptance and adoption) (Q13). Each question consists of several sub-questions (Q9: 4, Q10: 3, Q11: 4, Q12: 5, Q13: 6). The complete set of questions is provided in the Appendix.

The survey questions were developed based on the review of the literature on algorithmic bias in healthcare. Major issues related to algorithmic bias (e.g., distribution shifts, algorithm transparency) are covered in the questions. Questions about the intended behaviors were adapted from the survey instruments used for accessing technology acceptance and adoption in the information systems (IS) literature (Venkatesh et al., 2003).

A pilot study was conducted involving four participants, who were IS researchers with expertise in algorithmic bias and healthcare decision support systems. Questions were revised and modified for several rounds based on participants' feedback.

The participants of the full-scale survey were recruited from an executive MBA program offered at a business university in the northeastern USA. The executive MBA program was offered specifically to a cohort of employees in a world-class hospital based in the Greater Boston area. Students who were taking a healthcare analytics course in this program in spring 2024 were invited to participate in the survey. The survey was administered after covering various ML algorithms during the semester.

Among the 20 students who took the course, 19 responded to the invitation and participated in the survey. Among the 19 participants, the majority (n = 14) were female, and the rest (n = 5) are male. Students were from four age groups (i.e., 20-29, 30-39, 40-49, and 50+), and the number of participants in the groups was 4, 7, 4, 4, respectively. Their work experience ranged from 1-5 years (n = 3), 6-10 years (n = 6), 11-20 years (n = 7), to 20 or more years (n = 3). Their job roles included administrative (e.g., manager, director, team leader) (n = 4), clinical (e.g., physician, nurse, surgeon) (n = 8), technical (e.g., radiology) (n = 1), support (e.g., educator, analyst) (n = 4), and IT (e.g., developer, clinical system engineer) (n = 2). Regarding their experience with ML algorithms, all participants were familiar with decision tree and neural networks, which were covered in the course. Three participants also were familiar with other ML algorithms (e.g., k-nearest neighbor and naïve Bayes classifier).

## 3.2 Interviews

In a parallel study, we conducted semi-structured interviews with eight healthcare professionals in different positions/roles: doctors, nurses, therapists, and administrators overseeing medical IT. All recruited interviewees work in medical facilities in the northeastern USA; all of them were medically trained, although some were currently working in administrative positions, sometimes, in addition to their clinical work. In terms of type and area of practice, interviewees spanned a range from in-patient and outpatient care, emergency room (ER) care, primary care, to urgent care facilities.

Interview questions addressed perspective use of AI within EHRs and the issues surrounding ethics of using AI algorithms. The interviews lasting 30-45 minutes were conducted over Zoom, recorded, transcribed and analysed using inductive content analysis with the coding labels derived from the interview topic questions and emergent themes, refined in the process.

# 4 ANALYSIS AND RESULTS

## 4.1 Survey Results

Using the responses to each close-ended sub-questions as the dependent variable, we performed two-way ANOVA on three independent variables: age, gender, and job role. Since *years of experience in healthcare* and *years of experience in organization* are highly correlated with age, adding them to the ANOVA would cause the multicollinearity problem. The results from ANOVA using either experience variable produced similar results with those using age. Participants' *knowledge about ML algorithms* is homogeneous and does not show individual difference, since the cohort took the same analytics course where the algorithms were covered. Their *intended goals of using AI* correlate with their job roles (e.g., enhancing quality of care for clinical roles, reducing costs and improving operational productivity for administrative roles). As a result, these variables were not included in ANOVA. We summarize the ANOVA findings in the following.

### 4.1.1 Causes of algorithmic bias

The results show that the participants generally agree on the four statements about the major causes of algorithmic bias. In particular, they believe that bias in the *training data* can cause algorithmic bias (Q9.1 Likert scale mean = 4.6, S.D. = 0.51), and that the *distribution shift* of data, which occurs when an algorithm is trained on data from one hospital is used in another hospital, can also cause algorithmic bias (e.g., it may not work) (Q9.2 mean = 4.5, S.D. = 0.61). They are not as positive that a larger training dataset would necessarily mitigate bias (Q9.3 mean = 3.6, S.D. = 1.0). They tend to believe that bias may also come from the specific design of algorithms (Q9.4 mean = 3.9, S.D. = 0.88).

None of the three independent variables (age, gender, and job role) is significant in ANOVA across the four sub-questions. In other words, there is no gender, age, and role difference in the participants' opinions toward sources of algorithmic bias.

### 4.1.2 Fairness of outcomes

Participants strongly agree that if the outcome generated by an algorithm is biased against certain social groups, it is unfair for those groups (Q10.1 mean = 4.11, S.D. = 1.0). However, they disagree that protected demographic attributes (e.g., age, gender, race) of patients should be excluded in ML algorithms as a solution to prevent possible unfair outcomes (Q10.2 mean = 2.4, S.D. = 0.96). Instead, these attributes, which may potentially lead to unfair outcomes, can be used in some ML applications depending on the specific problem under study (Q10.3 mean = 4.4, S.D. = 0.60).

There is no age, gender, or role difference in the first two sub-questions in ANOVA. Only *age* is significant in the third sub-question, showing that the female participants (mean = 4.3) are slightly less inclined than the male participants (mean = 4.6) to allow ML algorithms to use demographical information of patients.

### 4.1.3 Assessment criteria and task characteristics

The responses to the sub-question regarding algorithm assessment are rather similar: the participants generally believe that ML algorithms should be assessed based on multiple criteria including performance, bias, and fairness (Q11.1 mean = 3.5, S.D. = 0.61), and that the prioritization of the criteria depends on the task characteristics (Q11.2 mean = 4.1, S.D. = 0.71). However, they are relatively neutral about whether algorithmic bias should be allowed in high- (e.g., fatal disease prediction) (Q11.3 mean = 3.5, S.D. = 1.0) or low-impact situations (e.g., patient satisfaction prediction) (Q11.4 mean = 3.5, S.D. = 0.84).

The ANOVA result shows that there is significant difference in age and job role for Q11.4. Among the four age groups, the 40-49 group gives significantly lower score to this question, indicating that they tend to disagree that algorithmic bias can be tolerated even in low-impact situations. In terms of job roles, clinicians also tend to disagree with this statement more than other participants with other roles.

### 4.1.4 Technology characteristics

Participants strongly believe that algorithm transparency (Q12.1 mean = 4.6, S.D. = 0.61) and explainability (Q12.3 mean = 4.3, S.D. = 0.73) are important, and they prefer algorithms with high transparency (Q12.2 mean = 4.3, S.D. = 0.75) and explainability (Q12.4 mean = 4.5, S.D. = 0.51) over those with lower transparency and explainability and similar performance. However, if one algorithm significantly outperforms another one, participants are less certain about whether they would choose the one with higher performance and lower transparency (Q12.5 mean = 3.2, S.D. = 1.1).

None of the independent variables is significant in ANOVA.

### 4.1.5 Intended behavioral responses

Participants' opinions are more divergent in terms of their intended behaviors (e.g. acceptance and adoption), given that ML algorithms vary in their performance (e.g., accuracy and error rate), bias, and fairness. Participants do not agree that ML algorithms should be outright abandoned even though their outcomes may have errors (e.g., false positives and false negatives) (Q13.1 mean = 2.7, S.D. = 1.0). However, they are more inclined to avoid using algorithms with biased (Q13.2 mean = 3.3, S.D. = 1.0) or unfair outcomes (Q13.3 mean = 4.0, S.D. = 1.1), but may still use the algorithm if its performance is high (Q13.5 mean = 3.4, S.D. = 1.1). They strongly agree that they will treat algorithm generated outcomes as mere recommendations and will rely on their own knowledge and experience to make final decisions (Q13.6 mean = 4.4, S.D. = 1.0).

The ANOVA identifies significant differences in job role, gender, and the interactions between role and age, and between gender and age for Q13.6. More specifically, the participants with administrative roles (mean = 3.25) are less likely to agree with the statement than other roles including clinical, technical, IT, and support (mean = 4.7).

## 4.2 Interview Results

In this section we summarize the interviewee responses regarding issues surrounding the use of AI and algorithmic bias. Full analysis of the interview data is beyond the scope of this paper. Here we present points relevant to our research questions, illustrated with quotes from the interviews.

### 4.2.1 Causes and consequences of errors and algorithmic bias

All interviewees saw the potential associated with the application of AI/ML in medical practice, while recognizing that the technology may not have reached the level needed to be used in clinical settings. Algorithm generated recommendations are part of the usual workflow for some practitioners. Practitioners referred to system generated, typically rule-based scoring of risks, such as risk of patient experiencing sepsis, falling, or heart attack. Interviewees mentioned that while they use such information, they typically combine it with other assessments instead of following it blindly. They emphasized that transparency in the way the scores are generated (typically, rule-based) is important to them and they realize that the scoring is not flawless. Among the sources of concern for potential inaccuracy and bias in the automatically generated recommendations, they mentioned issues of distribution shift, model bias, and the validity of the knowledge base used by algorithms to make the assessment. Concerns regarding the knowledge base include patient information that cannot be easily put into the written form or easily found, as well as the validity of literature.

*(INT_7) Mostly, we have a very diverse patient population and if the model has been designed and tested on a population that is very different from our population, certainly bias can be introduced in that way.*

One practitioner recognized the failure of a newly implemented ML system to account for the differences in the environment for recommending whether to discharge a patient to a rehab facility or to their home:

*(INT_2) One of the things that we're running into is the discharge tool that we're using doesn't take into account a person's home environment. So you know it might say that a patient can go home not realizing there's 30 steps to get into their home. But again, in Massachusetts, our architecture is very different than you get in Florida, so if this was a tool developed in Florida where a lot of the houses there are one level, no steps to get in, versus Massachusetts where you're dealing with architecture from the 17 - 18 hundreds in some cases.*

An emergency care physician noted the danger associated with over-reliance on tools (i.e., the automation bias), as an additional potential negative consequence of using inaccurate or biased algorithms:

*(INT_0) So I just worry that there are subtleties in the human condition that AI may not be able to analyze and give us that information. And we may become too reliant on it.*

Physicians emphasized that in the end they rely on their own judgement, as stated, for example, by this emergency care physician describing the use of patient acuity score:

*(INT_3) We use that as a guide, but we don't rely on it.*

A different concern is expressed by a family physician, who is also a chief medical informatics officer (CMIO), regarding AI potentially proposing case-relevant literature:

*(INT_5) And then the, you know, the concerns about bias as well. Kind of knowing that the data that the AI is using is unbiased data. You mentioned the AI looking at the latest literature and presenting that data to the clinician. There's a lot of junk in the literature now as you know, there are journals that are really not going through the peer review process, so just because it's the latest data doesn't mean that it's trustworthy. So, you know, I'd have concerns about bias as well. Those are all challenges. I'm sure there are others too.*

### 4.2.2 Assessment and response to bias

Approaches to addressing bias and inaccuracies in models were expressed by two professionals, who are both involved in assessment and implementation of ML-based tools. Both emphasized working closely with vendors to understand the parameters of the black-box models and data they were trained on.

Detecting and addressing bias issues before a tool is put into clinical use is a major concern of a chief medical informatics officer, who listed bias as one of the many evaluation parameters for the ML-based tools. In this person's opinion, there is a wide range of the level of physicians' awareness of the strengths and pitfalls of ML-based tools:

*(INT_7) Not many people are sort of thinking about how do we asses for bias? Is the tool safe to use? All those kinds of things…. And that's my job. And that's why we're not implementing a lot of tools right now that would provide clinical recommendations, at least.*

A more optimistic view on dealing with some known biases was expressed by this provider:

*(INT_2) I think if it's a known bias, that people know about, you're gonna automatically correct for that.*

A CMIO points out that the significance of different factors used in model accuracy evaluation depends on the specific task, for example, age may be more important than race in some settings:

*(INT_7) So with, something for radiology, I'd be thinking about age because structure is what radiology looks on, changes with age... But in other things, you certainly want to make sure that you, are looking at race and ethnicity a lot more...depending on the model you're looking at, you may be particularly cognizant of certain demographic characteristics that you want to make sure are included in in the study populations.*

Retrospective data evaluation is used to assess the impact of the distribution shift:

*(INT_7) We are more likely to feel comfortable with the model if we're going to test it on retrospective data with our population. And see that it performs well and if there is a plan for repeated validations going forward, then if there's a proposal to put it in with our population and just move forward prospectively.*

### 4.2.3 Technology adoption

Regarding adoption and use, transparency of the reasons for presented recommendations as well as the data used in model training was mentioned as a desirable factor by many interviewees. For the proprietary models, vendors do not necessarily disclose what data their model is trained on right now, although regulations, such as HTI 1 (US Department of Health and Human Services 2024), are being put in place requiring more disclosure:

*(INT_7) But going forward, they will have to be a lot more transparent with sharing a lot of information, not their entire, sort of, secret sauce, but a lot of information about their algorithm. But right now we work very closely with them, try to gain as much information as possible. But yeah, do we don't have, we don't have a lot of that information and so, you know, it's hard.*

Doctors and nurses emphasized that an ML-based clinical decision-making support tool's recommendation should be one of the points of information for them to consider. Many expressed the need for algorithm's transparency regarding the basis for the recommendation. The following quote describes the sentiment expressed by many interviewees:

*(INT_5) There's always the concern of the black box, you know, you don't know how the AI is arriving at these decisions. So transparency is definitely a limitation. I would love AI to make suggestions and not make decisions. So like, you know, I always want to make sure that there's some human reviewing everything, so you know can't make it completely automatic.*

Our interviewees also expressed many suggestions regarding the tasks for which they see AI/ML being useful, but this information is beyond the scope of this paper.

## 5 DISCUSSION

Recent advancements in AI and ML technologies are set to revolutionize healthcare and medical practice. However, the threat that the algorithmic bias inherent in many ML applications will amplify existing social biases and cause significant harm to the fair and safe medical care is real. Algorithmic bias requires focused attention by all stakeholders: developers of algorithms and healthcare systems, physicians, patients, insurance and other payers, and regulatory institutions. In our study, we conducted an initial investigation of perceptions of healthcare professionals regarding algorithmic bias, its sources, and their intended behaviors in responding to the algorithmic bias and fairness issues.

Healthcare professionals participating in the study are well aware of the issues of algorithmic bias, its relationship to the model design and training data, and the dangers of proliferating the unfair treatment through unguarded use of biased models. Those professionals who have experienced working with or evaluating ML models for clinical decision making have experienced dealing with issues of model bias, distribution shift, imbalanced training data, and non-transparency of recommendations. The prevailing attitude of surveyed and interviewed healthcare personnel, both in patient-facing and administrative roles, is that algorithmic recommendations should be treated as a point of information, with the physician's judgement applied as the deciding factor. Assessment of ML models for performance, safety, and fairness is a major concern for medical informatics officers and physicians involved in evaluating and implementing new technology. In assessing the model suitability for clinical practice, practitioners favor a differentiated approach to the inclusion of a variety of demographic and socio-economic factors in the training data. The differentiation is based on the specific task and domain of application of the model (e.g. radiology vs suicide prevention). Regarding the tolerance of bias and its importance compared to model performance, practitioners also have a differentiated approach based on the task. The approach to weighing performance versus potential bias does not depend on the importance of the decision (high-impact, fatal disease prediction, or low impact situation).

Practitioners strongly prefer technology that exhibits transparency and explainability of decisions, although they are less concerned with transparency for high-performing algorithms. In terms of the roles, administrators are less likely to agree to adopt flawed or biased algorithms into the practice. Administrators working on assessing and implementing ML-based technologies stress dialog with the developers, especially regarding the population characteristics of the training data, as a way to achieve satisfactory transparency, performance and fairness results from applying the ML models. Healthcare workers recognize that the different stakeholders have responsibilities in ensuring fairness and safety in the use of ML-based tools.

### 5.1 Implications for Research and Practice

Our study has several implications for both research and practice related to algorithmic bias. First, based on the theoretical framework on algorithmic bias (Kordzadeh & Ghasemaghaei, 2022), our study explores the relationships between the key constructs in the framework: algorithmic bias, perceived fairness, and intended behavioral responses, in the context of healthcare. Our findings show that algorithmic bias may have a negative impact on perceived fairness, which will further affect users' decisions for accepting or adopting ML algorithms. Second, using both survey and interview methods, we have gathered empirical evidence that different stakeholders have varying opinions toward algorithmic bias and fairness, which depend on several factors including the stakeholders' job roles, their perspectives, the particular task, and the technical characteristics of the algorithms and systems. Third, the findings from this study suggest that research on algorithmic bias in healthcare should focus on approaches to developing transparent solutions and communicating the known uncertainty in the model recommendations, including threats to model fairness, in clinical setting. Specifically, given the prominence of the distribution shift, as a source of bias, researchers should develop robust methodologies for assessing the distribution shift, as well as mitigating or overcoming it. It is important to study the impact of specific demographic and socio-economic factors for algorithm fairness for the specific domains of application, as it was noted by study participants.

Our findings also suggest that the practice of algorithmic bias mitigation should take serious consideration of the particular tasks and contexts, and that algorithms should be assessed and selected based on multiple performance criteria and ethical principles. Sometimes, it is necessary to prioritize these criteria and principles depending on the specific applications under study and the perspectives of stakeholders involved in the development of healthcare information systems. It is still a long journey to fully address algorithmic bias and ensure fairness and equity in healthcare.

### 5.2 Limitations

Limitations of this study include the small sample of surveyed and interviewed professionals as well as their limited geographic diversity. In recruiting for this initial study, we did not consider the race, nor the socio-economic status and other demographic characteristics of the patient population faced by the participants in their practice. A greater sample size would have enabled us to also stratify the participants by the level of knowledge of AI/ML in general.

## 6 CONCLUDING REMARKS

This research seeks to explore opinions of healthcare professionals facing rapid advancements of AI and ML in healthcare and the associated algorithmic bias and fairness issues. Our future research will extend to other important healthcare stakeholders. In particular, it will be an interesting research question to investigate how patients with different backgrounds, health conditions, and socioeconomic status view the prospects of AI in healthcare and the resulting ethical implications.

## ACKNOWLEDGEMENTS

We thank our study participants for their time and invaluable input.

# APPENDIX: SURVEY QUESTIONS

Q1. What is your gender?
- Male; Female; Prefer not to answer

Q2. What is your age?
- 20-29; 30-39; 40-49; 50+

Q3. What is your job role/function (check all that apply)?
- Clinical (e.g., physician, nurse, surgeon)
- Administrative (e.g., director, manager, team leader)
- Technical (e.g., radiologist)
- Support (e.g., educator, trainer, analyst)
- IT (e.g., developer, system engineer)
- Other, please specify ___________

Q4. How long have you been working at your hospital or organization?
- Less than one year; 1-2 years; 3-5 years; 6-9 years; 10-15 years; 15+ years

Q5. How long have you been working in the healthcare sector?
- Less than one year; 1-2 years; 3-5 years; 6-9 years; 10-15 years; 15+ years

Q6. Which of the following analytics and machine learning algorithms are you familiar with (check all that apply)?
- Regression (linear and logistic)
- Decision tree
- Neural networks
- Support vector machine
- K-nearest neighbor
- Naïve Bayes classifier
- Random Forest
- Other, please specify __________

Q7. Which will be the primary goal you wish to achieve with the use of AI (select the most applicable options)?
- To improve patient care quality
- To increase patient satisfaction
- To reduce cost
- To reduce errors
- To improve the performance of my organization
- To conduct research and publish papers
- Other, please specify__________

Q8. In general, what is your opinion about using AI technology in healthcare (check all that apply)?

- I think that AI has great potential to help improve my productivity.
- I think that AI has great potential to help improve healthcare quality and performance.
- I think the use of AI in healthcare may pose a tremendous amount of risks (e.g., misdiagnosis) onto patients.
- I think AI is a threat to healthcare professionals' job opportunity.
- I am concerned that the use of AI may cause privacy breaches of patient data.
- I think the adoption of AI technology may cause healthcare costs to increase.
- Other, please specify____________

In this study, we focus on a subset of AI techniques that use supervised machine learning algorithms to make decisions (e.g., disease diagnosis, physician referrals). In other words, the outcomes of the algorithms are classification labels (e.g., positive vs. negative). Without special notice, "algorithms" in the following statements refer to classification algorithms.

Q9. Please rate how much you agree (or disagree) with each of the following statements about the sources of algorithmic bias.

- Q9.1 The outcome of an algorithm is likely to be biased if the training data are biased.
- Q9.2 Algorithms trained on data from one hospital may not necessarily perform well when used in a different hospital.
- Q9.3 The larger the training dataset, the less likely an algorithm is biased.
- Q9.4 Algorithmic bias may result from algorithm design (e.g., the impurity measure used in decision trees).

Q10. Please rate how much you agree (or disagree) with each of the following statements about the fairness of the outcome produced by algorithms.

- Q10.1 If an algorithm's outcome is biased against certain social groups, it is unfair for those groups.
- Q10.2 To mitigate possible algorithmic bias, individual characteristics (e.g., race, gender, and age) should be excluded from all healthcare applications involving machine learning algorithms.
- Q10.3 Depending on the specific problems under study, such as those assessing risks for certain diseases (e.g., diabetes), individual characteristics (e.g., race, gender, and age) can be used in some applications involving machine learning algorithms.

Q11. Now imagine that you need to select algorithms to assist your decision making in your work. Please rate how much you agree (or disagree) with each of the following statements regarding how algorithms should be selected.

- Q11.1 Algorithms should be assessed based on multiple criteria such as performance, possible bias, fairness, etc.
- Q11.2 I would prioritize algorithm performance and bias differently depending on the specific situations.
- Q11.3 In high-impact situations (e.g., fatal disease prediction), algorithmic bias should not be allowed.
- Q11.4 In relatively low-impact situations (e.g., patient satisfaction prediction), algorithmic bias may be tolerated to a certain degree.

Q12. Please rate how much you agree (or disagree) with each of the following statements regarding algorithm transparency and explainability.

- Q12.1 Algorithm transparency is important for me to assess algorithmic bias.
- Q12.2 If two algorithms perform similarly, I prefer to use algorithms with high transparency (e.g., decision tree) over algorithms with low transparency (e.g., neural networks).
- Q12.3 Algorithmic explainability is important for me to assess algorithmic bias.
- Q12.4 If two algorithms perform similarly, I prefer to use algorithms with high explainability (e.g., decision tree) over algorithms with low explainability (e.g., neural networks).
- Q12.5 If one algorithm performs significantly better than another algorithm, I prefer to use the one with high performance even if its transparency or explainability is worse than the other algorithm.

Q13. Please rate how much you agree (disagree) with each of the following statements about how you treat algorithmic bias.

- Q13.1 If an algorithm's outcome has errors (e.g., false positives and false negatives), I will NOT use that algorithm to assist my decision making.
- Q13.2 If an algorithm's outcome is biased, I will NOT use that algorithm to assist my decision making.
- Q13.3 If an algorithm's outcome leads to unfair resource allocation among different social groups, I will NOT use that algorithm to assist my decision making.
- Q13.4 If the outcome of an algorithm is completely unbiased but its performance is low, I may still use the algorithm depending on the problem under study.
- Q13.5 If the outcome of an algorithm is biased but its performance is high, I may still use the algorithm depending on the problem under study.
- Q13.6 Being aware that algorithms may have errors and bias, I may still use algorithms but will treat the outcomes only as recommendations and rely on my own knowledge and experience to make the final decisions.